\documentclass{article}
\usepackage{spconf,amsmath,graphicx}

\usepackage{hyperref}       % hyperlinks
\hypersetup{
    colorlinks,
    linkcolor={red!90!black},
    citecolor={green!90!black},
    urlcolor={blue!30!black}
}
\usepackage{url}            % simple URL typesetting
\usepackage{booktabs}       % professional-quality tables
\usepackage{amsfonts}       % blackboard math symbols
\usepackage{xcolor}         % colors
\usepackage{amssymb}
\usepackage{makecell}
\newcommand{\etal}{\textit{et al}.}
\newcommand{\ie}{\textit{i}.\textit{e}.}
\newcommand{\eg}{\textit{e}.\textit{g}.}

\usepackage{pifont}
\newcommand{\cmark}{\ding{51}}
\newcommand{\xmark}{\ding{55}}
\usepackage{arydshln}
\usepackage{wrapfig}
\usepackage{multirow}

\DeclareMathOperator*{\argmax}{argmax}

\usepackage[
backend=biber,
style=ieee,
citestyle=numeric-comp,
maxbibnames=3,
maxcitenames=3,
doi=false,isbn=false,url=false,eprint=false
]{biblatex}
\defbibheading{bibliography}{\begin{center}\textbf{5. REFERENCES}\end{center}}

\title{Towards Practical and Efficient Image-to-Speech Captioning \\ with Vision-Language Pre-training and Multi-modal Tokens}
\name{Minsu Kim$^1$, Jeongsoo Choi$^1$, Soumi Maiti$^2$, Jeong Hun Yeo$^1$,  Shinji Watanabe$^2$, Yong Man Ro$^{1*}$\thanks{*Corresponding Author.}}
\address{$^1$Integrated Vision and Language Lab, KAIST, South Korea\\
$^2$Language Technologies Institute, Carnegie Mellon University, USA\\
\small{\texttt{\{ms.k,jeongsoo.choi,sedne246,ymro\}@kaist.ac.kr}, \,\,\texttt{\{smaiti,swatanab\}@andrew.cmu.edu}}}
\begin{document}
\ninept
\maketitle
\begin{abstract}
In this paper, we propose methods to build a powerful and efficient Image-to-Speech captioning (Im2Sp) model. To this end, we start with importing the rich knowledge related to image comprehension and language modeling from a large-scale pre-trained vision-language model into Im2Sp. We set the output of the proposed Im2Sp as discretized speech units, \ie, the quantized speech features of a self-supervised speech model. The speech units mainly contain linguistic information while suppressing other characteristics of speech. This allows us to incorporate the language modeling capability of the pre-trained vision-language model into the spoken language modeling of Im2Sp. With the vision-language pre-training strategy, we set new state-of-the-art Im2Sp performances on two widely used benchmark databases, COCO and Flickr8k. Then, we further improve the efficiency of the Im2Sp model. Similar to the speech unit case, we convert the original image into image units, which are derived through vector quantization of the raw image. With these image units, we can drastically reduce the required data storage for saving image data to just 0.8\% when compared to the original image data in terms of bits. Demo page: \href{http://bit.ly/3Z9T6LJ}{bit.ly/3Z9T6LJ}.
\end{abstract}
\section{INTRODUCTION}
Directly synthesizing a speech description for an image holds substantial promise in enhancing people's daily experiences. By narrating traffic signs on roads through generated speech descriptions, individuals with visual impairments can gain a comprehensive understanding of their immediate surroundings and route. Thereby, they can make informed decisions for their safe journey. Moreover, the capability to audibly check the image messages, even while engaged in activities like driving, can positively impact our daily routines. This Image-to-Speech captioning (Im2Sp) technology \cite{hasegawa2017image2speech} can be viewed as an audio counterpart of image captioning \cite{xu2015show} that predicts textual sentences describing input images. Despite the potential benefits of Im2Sp, the technology has not been well-addressed compared to image captioning. Different from text-based image captioning, developing an end-to-end Im2Sp model is regarded as a challenging problem, due to the weak supervision of speech regression in comprehending the visual input \cite{van2017neuraldiscrete,kim2023liptospeech}. As speech contains not only linguistic information but also various irrelevant factors (\eg, speaker characteristics, duration, noises) to the input image, guiding the model with regression criteria that force to produce speech features (\eg, Mel-spectrogram) similar to that of ground-truth speech, may prevent the model from focusing on the image content \cite{wang2021synthesizing,kim2023liptospeech}.

These days, discretized speech unit \cite{lakhotia2021generative} has drawn big attention with its significant potential in diverse tasks such as speech-to-speech translation \cite{inaguma2022unity,popuri2022enhanced,kim2023many}, spoken language understanding \cite{sicherman2023analysing,nguyen2023generative}, speech synthesis \cite{hayashi2020discretalk,choi2023intelligible}, and speech recognition \cite{hsu2021hubert,chang2023exploration,kim2023lip}. The speech units can be obtained by quantizing speech features derived from self-supervised speech models. Since they are discrete and can be generated to exclusively encapsulate linguistic factors (\ie, phoneme) \cite{lakhotia2021generative,sicherman2023analysing,kim2023many}, speech units can serve as pseudo-text. By utilizing the pseudo-text characteristics of the speech unit, one can build an end-to-end Im2Sp model by guiding the model with strong discrete supervision (\eg, classification) instead of using a regression criterion. Nevertheless, the performance of the Im2Sp model remains notably lower than that of image captioning, making it inadequate for practical real-world utilization. Since acquiring paired data of images and human spoken speech is challenging, the limited training data makes it difficult for models to learn how to comprehend images and convert them into speech descriptions. As jointly understanding the image and speech is one of the key elements in developing multi-modal language technologies \cite{chrupala2017representations,shih2023speechclip,hong2023watch}, it is important to devise an approach for associating the image and speech even when faced with limited image-speech paired data.

In this paper, we focus on improving the performance of an end-to-end Im2Sp model. To this end, we investigate whether the rich knowledge of image understanding and language generation of a large-scale pre-trained vision-language model \cite{radford2021clip,wang2022git} can be transferred to Im2Sp. Then, we show that even if the vision-language model is pre-trained with image-text modalities instead of speech, we can significantly improve the performance of Im2Sp by incorporating its pre-trained knowledge. Furthermore, we explore how we can enhance the efficiency of the Im2Sp model. Similar to the speech unit case, we quantize the input image into image units. Concretely, we tokenize the input image into image units by applying Vector Quantization (VQ) technique of ViT-VQGAN \cite{esser2021vqgan,yu2021VIM}. With the tokenized inputs, our Im2Sp problem becomes a translation between multi-modal tokens like language translation \cite{kim2023many}. In this setup, both the input and output are discrete, enabling efficient model training and economic data storage management. The image units reduce the required bits more than 100 times compared to the original raw image. We show that with the vision-language pre-training strategy, we can still achieve reasonable Im2Sp performances while effectively reducing the required data storage and computational memory costs.

The major contributions of this paper can be summarized as follows: 1) This is the first work exploring vision-language pre-training in Im2Sp. By employing the vision-language pre-trained image encoder and text decoder in our Im2Sp framework, we achieve state-of-the-art performances, demonstrating a significant performance margin compared to previous methods on two popular benchmark databases, COCO \cite{lin2014coco} and Flickr8k \cite{hodosh2013flickr8k}. 2) This is the first work investigating the image token-to-speech token translation framework with NLP-like processing of multi-modality, which can greatly reduce the required data storage. 3) Through comprehensive experiments including caption quality evaluations, human subjective evaluation, and state-of-the-art neural MOS evaluation \cite{lo2019mosnet,maiti2023speechlmscore}, we show the proposed Im2Sp model can generate natural speech with having the correct description for input images.

\vspace{-0.2cm}
\section{METHOD}
\begin{figure}[t]
	\begin{minipage}[b]{1.0\linewidth}
		\centering
		\centerline{\includegraphics[width=9.2cm]{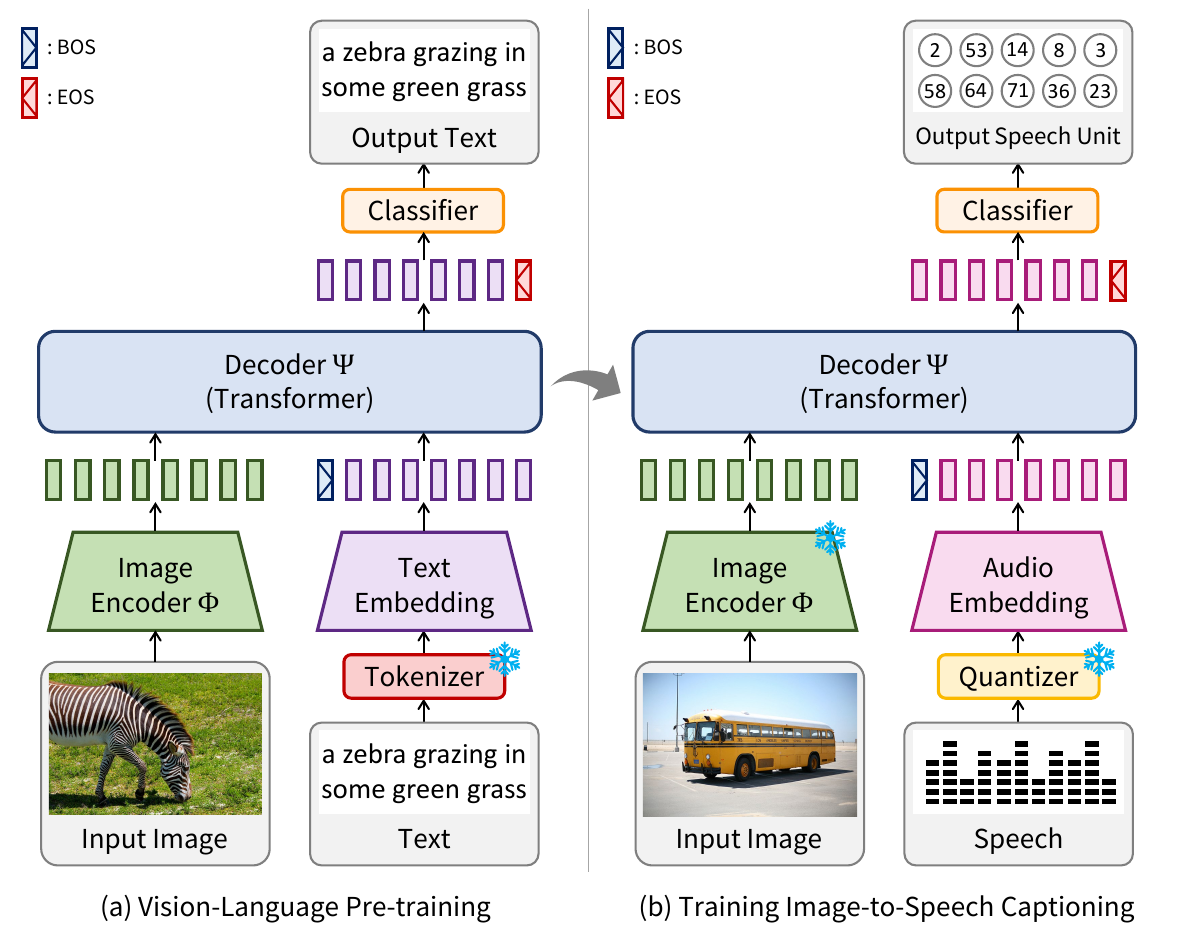}}
	\end{minipage}
	\vspace{-0.8cm}
	\caption{Illustration of the proposed image-to-speech captioning framework. (a) By employing the vision-language pre-training strategy, (b) we can bring the learned knowledge of image comprehension and language generation into our image-to-speech captioning.}
	\label{fig:1}
	\vspace{-0.4cm}
\end{figure}

\vspace{-0.1cm}
Fig. \ref{fig:1} shows the proposed Im2Sp framework. Let $x\in\mathbb{R}^{H\times W\times C}$ be an input image and $y\in\mathbb{R}^{T}$ be the ground-truth speech caption with a sample rate of 16kHz. Here, $H$, $W$, and $C$ represent the image size of height, width, and channel, respectively, and $T$ represents the length of the waveform. The main objective of our learning problem is to translate the input image $x$ into speech $y$ that correctly describes the image content. 
To improve the performance of Im2Sp, we propose leveraging the knowledge of a pre-trained model trained on large-scale image-text data. Moreover, we improve the efficiency of the Im2Sp model by introducing multi-modal tokens. The details of the proposed method are described in the following subsections.
%For our target speech features, we utilize speech units, which are the quantized speech features derived from the self-supervised speech model, HuBERT \cite{hsu2021hubert}. To maximize the capability of our Image-to-Speech captioning (Im2Sp) model in comprehending the input image and associating it with language, we explore vision-language pre-training strategies in our Im2Sp training. Finally, we develop an efficient Im2Sp model whose required data storage is reduced more than 100 times, by quantizing input raw images into image units.

\vspace{-0.1cm}
\subsection{Speech Unit Extraction}
Previous Im2Sp methods \cite{hsu2021text,wang2021synthesizing,effendi2021end} showed that by utilizing discovered discrete acoustic units instead of directly predicting continuous speech features (\eg, Mel-spectrogram), we can improve the performance of Im2Sp. This is because by extracting the discrete acoustic units from the speech, we can focus more on the linguistic modeling of speech while suppressing the other factors in the speech \cite{lakhotia2021generative,sicherman2023analysing}.

Different from the previous works \cite{hsu2021text,effendi2021end} that utilize discrete acoustic units derived from Mel-spectrogram such as the codebook of VQ-VAE, we utilize speech units, discovered from the recent self-supervised speech model, HuBERT \cite{hsu2021hubert}. Hence, we eliminate the need for complex processes involving predicting the Mel-spectrogram from discrete acoustic units and converting the raw waveform from the predicted Mel-spectrogram, as required by previous methods. Instead, we can directly convert the waveform from the speech units by utilizing a speech unit-based vocoder \cite{polyak2021speech,kong2020hifi}, with even more natural speech sound. Specifically, we extract speech features using a pre-trained HuBERT \cite{hsu2021hubert} and perform K-means clustering to obtain the discretized units, following \cite{lakhotia2021generative}. Then, we remove sequential repetitions of the units and finally obtain our speech units $u\in\{1,\dots,N_u\}^S$ which will be used for the proposed Im2Sp. Here, $N_u$ and $S$ represent the token size and length of speech units, respectively. As HuBERT downsamples the raw audio $y$ by a factor of 320, our speech units $u$ have a much lower frame rate than the raw audio (\ie, $S < T/320$). 

%------------------------------------ Figure 1
%#################################################

%##################################################

\vspace{-0.1cm}
\subsection{Image-to-Speech with Vision-Language Pre-training}
Fig. \ref{fig:1}b shows the overall architecture of the proposed Im2Sp model. It is mainly composed of an image encoder $\Phi$ and a speech decoder $\Psi$. The image encoder is designed with Vision Transformer (ViT) \cite{dosovitskiy2020vit} which is showing promising results in diverse vision tasks \cite{han2022survey}. When an input image $x$ is given, the image encoder $\Phi$ extracts the visual features $f_v$ by downsampling the spatial size as follows, $f_v = \Phi(x) \in \mathbb{R}^{(H/P * W/P + 1) \times D}$, where $P$ represents the patch size of ViT, $D$ represents the embedding dimension, and the additional 1 dimension (\ie, +1) comes from the attached CLS token. By treating the flattened spatial region of $f_v$ as a sequence, it is employed as a visual condition for the speech decoder $\Psi$. Therefore, the speech decoder $\Psi$ can generate the speech units $u$ describing the conditioned visual features $f_v$. After the visual features $f_v$, an embedding of BOS (Beginning of Sequence) token is attached and the speech decoder predicts the speech units $u$ in an autoregressive manner until EOS (End of Sequence) is predicted. The objective function of the proposed Im2Sp can be represented as follows,
\begin{equation}
\label{eq:1}
\argmax_\theta \sum_{k=1}^S \log{p(u^k|u^{<k},x;\theta)},
\end{equation}
where $u^k$ represents the current prediction, $u^{<k}$ represents the previous prediction, and $\theta$ is the model parameters including image encoder, text decoder, and embedding layers for speech units.

Motivated by the recent progress in vision-language pre-training (Fig. \ref{fig:1}a) \cite{radford2021clip,wang2022git}, we try to bring the image understanding knowledge and language generation knowledge of the large-scale pre-trained vision-language model into our Im2Sp model. Hence, we can alleviate the limitation in the Im2Sp task, where there is relatively limited availability of paired image and human spoken speech compared to the abundance of image-text paired data. Specifically, both the image encoder and the speech decoder are initialized from a pre-trained vision-language model, GiT \cite{wang2022git}. GiT is pre-trained with text generation from images, thus the model knows how the image can be comprehended and can be described in language. Please note that the weight of the speech decoder is initialized with the text decoder of GiT. As the speech units mainly hold linguistic information \cite{chang2023exploration,kim2023many}, we can transfer the language modeling ability of the pre-trained text decoder of the vision-language model into our spoken language generation \cite{kim2023many}. The knowledge transferring from the vision-language model into the Im2Sp model is shown in Fig. \ref{fig:1}.

\vspace{-0.1cm}
\subsection{Efficient Image-to-Speech Captioning with Image Units}
\label{sec:2.3}
%------------------------------------ Figure 2
%#################################################
\begin{figure}[t]
	\begin{minipage}[b]{1.0\linewidth}
		\centering
		\centerline{\includegraphics[width=7.3cm]{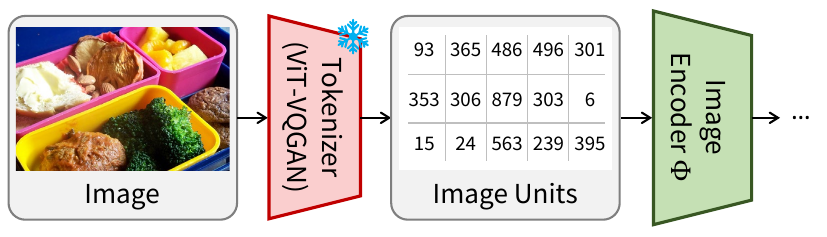}}
	\end{minipage}
	\vspace{-0.8cm}
	\caption{The extraction of image units by using vector quantization. Extracted image units are utilized for inputs instead of raw images.}
	\label{fig:2}
	\vspace{-0.4cm}
\end{figure}
%##################################################

Multi-modal processing systems, especially utilizing visual and audio modalities, require much more data storage and computational memory costs than text-only systems. This is why training large-scale multi-modal speech processing systems is significantly more challenging than NLP systems, with most of the development taking place in the industry. These days, \cite{mnih2014neural,van2017neuraldiscrete} showed that we can represent the image with compressed discrete representations while maintaining its content by applying Vector Quantization (VQ) to the continuous image features. To assess the feasibility of creating efficient multi-modal processing systems, we investigate the Im2Sp system working with quantized image representations, the image units. Therefore, our system now takes discrete image tokens as input and generates discrete speech tokens as output, resembling the operation of an NLP system that works with discrete text input and output \cite{kim2023many}.

To this end, we employ a pre-trained image vector quantizer of ViT-VQGAN \cite{esser2021vqgan,yu2021VIM}, as shown in Fig. \ref{fig:2}. The quantizer tokenizes the input image $x$ into image units $i\in\{1,\dots,N_i\}^{H/8\times W/8}$ by downsampling its spatial size with a factor of 8. The token size of image units $N_i$ is 8,192 (13 bits). Then, with the image units, we train the Im2Sp model with the aforementioned vision-language pre-training strategy. Therefore, we first train an image-to-text system and then transfer the knowledge into the Im2Sp model. To employ image units as inputs for the image encoder, we follow SEiT \cite{park2023seit} and utilize Stem-Adpator to handle the different input sizes. As the system purely works with discrete inputs and outputs, the required data size can be greatly reduced. We compare the bit size of different input \cite{park2023seit} and output \cite{chang2023exploration} representations in Table \ref{table:1}. By utilizing image units, we can reduce the required bits to 0.8\% compared to the raw image. Moreover, by utilizing speech units at the output side, we only require 0.2\% bits compared to raw waveform (based on 16bit, 16kHz audio) or Mel-spectrogram (based on 100 FPS and 80 mel-spectral dimensions). As we remove the repetition of speech units, we can further reduce the data size, similar to that reported in \cite{chang2023exploration}. As a result, we can significantly shrink the amount of data storage and GPU memory needed for training the model for both input and output parts. This makes it much easier to scale multi-modal processing systems to large-scale training.

\vspace{-0.2cm}
\section{EXPERIMENTS}
\vspace{-0.1cm}
\subsection{Dataset}
We utilize two Im2Sp databases, Flickr8kAudio \cite{harwath2015flickr8kaudio} and SpokenCOCO \cite{hsu2021text}. For both datasets, Karpathy split \cite{karpathy2015deep} is used. Flickr8kAudio is a spoken version of Flickr8k \cite{hodosh2013flickr8k} recorded from 183 speakers. It consists of 6,000 images for training, and 1,000 images for validation and testing, respectively. Each image has 5 speech captions. SpokenCOCO is a spoken version of COCO-2014 captioning dataset \cite{lin2014coco} and is collected by recording the utterances from 2,532 speakers. It has 82,783 training images with 5,000 images for validation and testing, respectively. Five speech captions are provided for each image. For training, the two datasets are utilized together following \cite{hsu2021text}. Then, the model is evaluated on each validation and test splits of COCO \cite{lin2014coco} and Flickr8k \cite{hodosh2013flickr8k}. For measuring the performance, we employ an off-the-shelf ASR model \cite{baevski2020wav2vec} to transcribe the generated speech. Then, we measure BLEU-4 \cite{papineni2002bleu}, METEOR \cite{denkowski2014meteor}, ROUGE \cite{lin2004rouge}, CIDEr \cite{vedantam2015cider}, and SPICE \cite{anderson2016spice}, which are the standard metrics in image captioning \cite{lin2014coco}, where all metrics indicate better performance with higher values.

%------------------------------------ Table 1
%#################################################
\begin{table}[t]
	\renewcommand{\arraystretch}{1.2}
	\renewcommand{\tabcolsep}{2.3mm}
\vspace{-0.2cm}
\caption{Data size (bits) comparisons according to different data types for image and audio modalities. Based on the image size of 224$\times$224, audio of 16kHz and 16bits, and Mel-spectrogram of 100 FPS and 80 filter banks. $L$ represents the time length of the audio.}
\vspace{0.1cm}
\centering
\resizebox{0.9\linewidth}{!}{
\begin{tabular}{cccc}
\Xhline{3\arrayrulewidth}
\textbf{Modality} & \textbf{Data Type} & \textbf{Data Size (bits)} & \textbf{Reduction Rate}\\ \hline
\multirow{2}{*}{\textbf{Image}} & Raw Image & 224 $\times$ 224 $\times$ 3 $\times$ 8 & 100\% \\
& Image Unit & 28 $\times$ 28 $\times$ 13  & 0.8\%\\ \hline
\multirow{3}{*}{\textbf{Audio}} & Raw Audio &  16000 $\times$ L $\times$ 16 & 100\% \\
& Mel-spectrogram & 100 $\times$ 80 $\times$ L $\times$ 32 & 100\% \\ 
& Speech Unit & ($<$50) $\times$ L $\times$ 8 & $<$0.2\%\\ 
\Xhline{3\arrayrulewidth}
\end{tabular}}
\label{table:1}
\vspace{-0.4cm}
\end{table}
%#################################################

%------------------------------------ Table 2
%#################################################
\begin{table*}[t]
	\renewcommand{\arraystretch}{1.2}
	\renewcommand{\tabcolsep}{1.4mm}
\vspace{-0.2cm}
\caption{Ablation study to confirm the effectiveness of vision-language pre-training in image-to-speech captioning.}
\vspace{0.05cm}
\centering
\resizebox{0.85\linewidth}{!}{
\begin{tabular}{cc ccccc ccccc}
\Xhline{3\arrayrulewidth}
\multicolumn{2}{c}{\textbf{Vision-Language Pre-training}} & \multicolumn{5}{c}{\textbf{Flickr8k}} & \multicolumn{5}{c}{\textbf{COCO}} \\ \cmidrule(l{2pt}r{2pt}){1-2} \cmidrule(l{2pt}r{2pt}){3-7} \cmidrule(l{2pt}r{2pt}){8-12}
\textbf{Image Encoder} & \textbf{Speech Decoder} & \textbf{BLEU-4} & \textbf{METEOR} & \textbf{ROUGE} & \textbf{CIDEr} & \textbf{SPICE} & \textbf{BLEU-4} & \textbf{METEOR} & \textbf{ROUGE} & \textbf{CIDEr} & \textbf{SPICE} \\ \cmidrule(l{2pt}r{2pt}){1-2} \cmidrule(l{2pt}r{2pt}){3-7} \cmidrule(l{2pt}r{2pt}){8-12}
\xmark & \xmark & 12.9 & 17.1 & 40.7 & 31.4 & 10.3 
& 17.4 & 19.1 & 44.0 & 50.5 & 12.2 \\
\cmark & \xmark & 17.7	& 20.6 & 45.9 & 45.8 & 14.0
& 20.9 & 21.3 & 46.3 & 64.7 & 15.1 \\
\cmark & \cmark & \textbf{20.6} & \textbf{22.0} & \textbf{48.4} & \textbf{53.6} & \textbf{15.8} 
& \textbf{25.9} & \textbf{23.8} & \textbf{50.4} & \textbf{81.1} & \textbf{17.5} \\
\Xhline{3\arrayrulewidth}
\end{tabular}}
\label{table:2}
\vspace{-0.4cm}
\end{table*}
%#################################################

%------------------------------------ Table 3
%#################################################
\begin{table*}[t]
	\renewcommand{\arraystretch}{1.4}
	\renewcommand{\tabcolsep}{0.8mm}
\vspace{-0.1cm}
\caption{Image-to-speech captioning performance comparisons on Flickr 8k and COCO. We also report the performance of image captioning and cascaded models for analysis purposes. We utilize an off-the-shelf TTS model \cite{kim2021conditional} for measuring the performance of cascaded models.}
\vspace{0.05cm}
\centering
\resizebox{0.85\linewidth}{!}{
\begin{tabular}{cc ccccc ccccc}
\Xhline{3\arrayrulewidth}
\multirow{2}{*}{\textbf{Modality}} & \multirow{2}{*}{\textbf{Methods}} & \multicolumn{5}{c}{\textbf{Flickr8k}} & \multicolumn{5}{c}{\textbf{COCO}} \\ \cmidrule(l{2pt}r{2pt}){3-7} \cmidrule(l{2pt}r{2pt}){8-12}
& & \textbf{BLEU-4} & \textbf{METEOR} & \textbf{ROUGE} & \textbf{CIDEr} & \textbf{SPICE} & \textbf{BLEU-4} & \textbf{METEOR} & \textbf{ROUGE} & \textbf{CIDEr} & \textbf{SPICE} \\ \cmidrule(l{2pt}r{2pt}){1-2} \cmidrule(l{2pt}r{2pt}){3-7} \cmidrule(l{2pt}r{2pt}){8-12}
\multirow{3}{*}{\makecell{Image captioning \\ (Image$\rightarrow$Text)}} & SAT \cite{xu2015show} & 21.3 & 20.3 & - & - & - & 24.3 & 23.9 & - & - & - \\
& Ours & 30.8 & 26.9 & 55.8 & 93.8 & 20.0 & 38.7 & 29.5 & 59.1 & 131.2 & 23.3 \\ \cdashline{2-12}
& Ours (Image Unit) & 23.4 & 22.0 & 48.9 & 63.3 & 15.4 & 29.9 & 25.2 & 52.8 & 97.4 & 18.6 \\ \cmidrule(l{2pt}r{2pt}){1-2} \cmidrule(l{2pt}r{2pt}){3-7} \cmidrule(l{2pt}r{2pt}){8-12}
\multirow{2}{*}{\makecell{Cascaded\\(Image$\rightarrow$Text \&\\Text$\rightarrow$Speech)}}
& Ours & 29.1 & 26.0 & 54.6 & 84.9 & 18.9 & 36.1 & 28.4 & 57.5 & 117.2 & 21.9 \\ \cdashline{2-12}
& Ours (Image Unit) & 22.3 & 21.3 & 48.0 & 57.7 & 14.6 & 28.2 & 24.3 & 51.6 & 87.4 & 17.5 \\ \cmidrule(l{2pt}r{2pt}){1-2} \cmidrule(l{2pt}r{2pt}){3-7} \cmidrule(l{2pt}r{2pt}){8-12}
\multirow{5}{*}{\makecell{\textbf{Image-to-Speech}\\\textbf{Captioning}\\\textbf{(Image$\rightarrow$Speech)}}}
& Wang \etal~\cite{wang2021synthesizing} & 3.5 & 11.3 & 23.2 & 8.0 & - & - & - & - & - & - \\
& SAT-FT-VQ3 \cite{hsu2021text} & 12.5 & 14.5 & 39.1 & 24.5 & 9.5 & 23.3 & 21.2 & 47.8 & 73.2 & 14.9 \\
& Effendi \etal~\cite{effendi2021end} & 14.8 & 17.4 & 32.9 & 45.8 & - & - & - & - & - & - \\
& \textbf{Ours} & \textbf{20.6} & \textbf{22.0} & \textbf{48.4} & \textbf{53.6} & \textbf{15.8} & \textbf{25.9} & \textbf{23.8} & \textbf{50.4} & \textbf{81.1} & \textbf{17.5} \\ \cdashline{2-12}
& \textbf{Ours (Image Unit)} & 16.7 & 19.6 & 44.2 & 41.2 & 13.1 & 20.1 & 21.4 & 46.4 & 64.0 & 15.0 \\
\Xhline{3\arrayrulewidth}
\end{tabular}}
\label{table:3}
\vspace{-0.5cm}
\end{table*}
%#################################################

\vspace{-0.1cm}
\subsection{Implementation Details}
Basically, our Im2Sp model has the similar architecture of GiT-large \cite{wang2022git} whose image encoder is ViT-large \cite{dosovitskiy2020vit} with a patch size of 14 (\ie, $P$=14) and decoder is composed of 6-layered transformers \cite{vaswani2017attention}. For the input image, we resize images to a size of 224$\times$224. For the speech unit extraction (Quantizer in Fig. \ref{fig:1}b), we use a pre-trained HuBERT-base model \cite{hsu2021hubert} and perform K-means clustering on features extracted at the 6th layer into 200 units (\ie, $N_u$=200), following \cite{nguyen2023generative}. To generate a waveform, we train a unit-based HiFi-GAN \cite{kong2020hifi,polyak2021speech} on LJSpeech \cite{ljspeech17}. For training the Im2Sp model, the image encoder and speech decoder are initialized from pre-trained GiT-large of \cite{wang2022git}. We freeze the image encoder, and only train the speech decoder and unit embedding layers, for 100k steps with a batch size of 64, a learning rate of $5e^{-5}$ with a warmup for 10k steps. Models are selected based on the BLEU score on the validation set. For training the image unit-based Im2Sp model, we first pre-train the image unit-based vision-language model on CC3M \cite{sharma2018cc3m}, SBU \cite{ordonez2011sbu}, COCO, and Flickr8k by initializing the image encoder with a pre-trained SEiT \cite{park2023seit}. The same text tokenizer with GiT is utilized. Then, the pre-trained image-text model is transferred into Im2Sp.

\vspace{-0.1cm}
\subsection{Experimental Results}
\quad\,\, {\bf Effectiveness of vision-language pre-training.}
To confirm the effectiveness of vision-language pre-training strategies, we train three variants of the Im2Sp model. 1) The baseline that does not utilize the strategy and just initializes the image encoder with a pre-trained image classifier \cite{dosovitskiy2020vit}, similar to \cite{hsu2021text}. 2) The model whose image encoder is initialized with the vision-language pre-trained model, CLIP \cite{radford2021clip}. 3) The proposed model that both image encoder and text decoder are initialized from the vision-language pre-trained model, GiT \cite{wang2022git}. The speech decoders for the first two models are randomly initialized and trained on Im2Sp datasets. Table \ref{table:2} shows the ablation results on Flickr8k and COCO. Without utilizing the vision-language pre-training, we achieve 12.9 and 17.4 BLEU scores on each database. By initializing the image encoder with pre-trained CLIP using image-text association, we can greatly improve the performances on all metrics from the baseline. Therefore, we can confirm that by utilizing a vision-language pre-trained image encoder instead of a simple image classifier, the model can better capture the language-associated semantics from input images. Next, when we additionally initialize the speech decoder with a pre-trained text decoder, we can further improve the performance. This is due to the fact that speech units mainly hold linguistic information and can be regarded as unified representations of speech and text \cite{kim2023many}, enabling the speech decoder to inherit the language model knowledge of the large-scale pre-trained text decoder.

{\bf Comparisons with the state-of-the-art methods.}
Table \ref{table:3} shows the evaluation results on Flickr8k and COCO databases. For analysis purposes, we also report the performance of image captioning and cascaded (\ie, image captioning \& text-to-speech) systems. Note that our text-based systems (\ie, image captioning and cascaded) are the models before transferred to Im2Sp, which are trained on over 3M image-text pairs \cite{wang2022git}. As the Im2Sp model is trained on 89K image-audio pairs, a direct comparison cannot be made between different modal systems. We highlight that even though the performance of the Im2Sp model is lower than the cascaded system, we still need to develop an end-to-end Im2Sp model for the following reasons. 1) More than 40\% of languages have no writing systems \cite{lee2022textless}, so the text-based model is not feasible for them. 2) We can reduce the inference time and maintenance costs compared to using two systems of image captioning and text-to-speech. Through continuous research efforts, we may achieve performance comparable to cascaded systems, much like what has been accomplished with end-to-end speech translation \cite{antonios2022findings}.

By comparing the performance of the proposed Im2Sp method with the previous state-of-the-art methods \cite{hsu2021text,effendi2021end}, we can confirm that the proposed method outperforms the previous methods with large gaps in all metrics. For example, the proposed Im2Sp achieves a 20.6 BLEU score on Flickr8k which outperforms the previous method \cite{effendi2021end} by 5.8 BLEU score. Furthermore, in contrast to previous methods that exhibited significantly lower performance than the popular image captioning system, SAT \cite{xu2015show}, the proposed Im2Sp model can now catch up with the performance of the text-based system (\ie, SAT). Please note that the works \cite{hsu2021text,effendi2021end} utilized ASR models trained on audio reconstructed from their audio features, hence some incorrect pronunciations are calibrated by the ASR model. In contrast, we achieve better performance by using an off-the-shelf ASR model \cite{baevski2020wav2vec}. We strongly recommend listening to the generated speech that is available on \href{http://bit.ly/3Z9T6LJ}{bit.ly/3Z9T6LJ}.

We also conduct Mean Opinion Score (MOS) tests, involving 15 participants who assessed 20 samples for each method. The subjects are asked to rate the naturalness of the generated speech and how correctly the generated speech describes the input image on a scale of 1 to 5. Moreover, we also report DNN-based MOS using MOSNet \cite{lo2019mosnet} and SpeechLMScore \cite{maiti2023speechlmscore}. The MOS comparison results are shown in Table \ref{table:4}. The results on both human and DNN-based metrics clearly show that the proposed Im2Sp method generates more natural sound with better descriptiveness than the previous method.

%------------------------------------ Table 4
%#################################################
\begin{table}[t]
	\renewcommand{\arraystretch}{1.3}
	\renewcommand{\tabcolsep}{1.2mm}
\vspace{-0.2cm}
\caption{Mean Opinion Score (MOS) comparisons with 95\% confidence interval, and Neural MOS scores on COCO.}
\vspace{0.05cm}
\centering
\resizebox{0.999\linewidth}{!}{
\begin{tabular}{ccccc}
\Xhline{3\arrayrulewidth}
\multirow{2}{*}{\textbf{Methods}} & \multicolumn{2}{c}{\textbf{Human Evaluation (MOS)}} & \multicolumn{2}{c}{\textbf{Neural MOS} \cite{lo2019mosnet,maiti2023speechlmscore}} \\ \cmidrule(l{2pt}r{2pt}){2-3} \cmidrule(l{2pt}r{2pt}){4-5}
& \textbf{Naturalness} & \textbf{Descriptiveness} & \textbf{MOSNet} $\uparrow$ & \textbf{SpeechLMScore} $\downarrow$ \\ \cmidrule(l{2pt}r{2pt}){1-3} \cmidrule(l{2pt}r{2pt}){4-5}
SAT-FT-VQ3 \cite{hsu2021text} & 2.870$_{\pm0.095}$ & 2.978$_{\pm0.131}$ & 4.12 & 4.25 \\
\textbf{Ours} & \textbf{4.275}$_{\pm0.086}$ & \textbf{3.968}$_{\pm0.108}$ & 4.26 & 4.17 \\ \hdashline
\textbf{Ours (Image Unit)} & 4.228$_{\pm0.089}$ & 3.725$_{\pm0.122}$ & \textbf{4.33} & \textbf{4.16} \\
\Xhline{3\arrayrulewidth}
\end{tabular}}
\label{table:4}      
\vspace{-0.6cm}
\end{table}
%#################################################

{\bf Performance of image unit-based system.}
The last row of Table \ref{table:3} shows the Im2Sp performance of the image unit-based system. We can find that there is a trade-off between efficiency and performance, similar to \cite{park2023seit}. However, we can achieve reasonable performances by achieving better performances than the previous state-of-the-art \cite{effendi2021end} on Flickr8k data. From the MOS test in Table \ref{table:4}, we find that we lose some descriptiveness when we use image units, but we can maintain the speech quality. Please note that with the unit-based Im2Sp, we can reduce a great amount of data storage and computation costs. The required bit size is reduced to 0.8\% and lower than 0.2\% for input and output, compared to original signals (Sec. \ref{sec:2.3}).

\vspace{-0.2cm}
\section{CONCLUSION}
\vspace{-0.1cm}
In this paper, we proposed a practical and efficient Image-to-Speech captioning (Im2Sp) method. We showed that even if speech is not utilized in vision-language pre-training, the knowledge of image comprehension and language modeling can be transferred into the Im2Sp model. Finally, by employing image units instead of raw images as inputs for our system, we showed that we can greatly reduce the data size (bits) while still achieving reasonable performances.

\printbibliography[heading=bibliography]

\end{document}